\documentclass[10pt,conference]{IEEEtran}

\IEEEoverridecommandlockouts

\usepackage{cite}
\usepackage{booktabs}
\usepackage{amsmath,amssymb,amsfonts}
\usepackage{graphicx}
\usepackage{textcomp}
\usepackage{multirow}        
\usepackage{url}
\usepackage{hyperref}
\hypersetup{hidelinks}

\usepackage{placeins}

\usepackage{caption}
\usepackage{subcaption}

\usepackage{balance}

\graphicspath{{./}{./figures/}{./images/}}
\DeclareGraphicsExtensions{.pdf,.png,.jpg,.jpeg}

\begin{document}

\title{Evo-DKD: Dual-Knowledge Decoding for Autonomous Ontology Evolution in Large Language Models\thanks{\copyright~2026 IEEE. Personal use of this material is permitted. Permission from IEEE must be obtained for all other uses, in any current or future media, including reprinting/republishing this material for advertising or promotional purposes, creating new collective works, for resale or redistribution to servers or lists, or reuse of any copyrighted component of this work in other works. Accepted for publication at the IEEE International Conference on Machine Learning and Applications (ICMLA) 2026.}}

\author{
\IEEEauthorblockN{Vishal Raman}
\IEEEauthorblockA{\textit{Independent Researcher}\\
Bethesda, USA\\
vishalraman613@gmail.com}
\and
\IEEEauthorblockN{Vijai Aravindh R}
\IEEEauthorblockA{\textit{Independent Researcher}\\
Chennai, India\\
vijairaman2003@gmail.com}
\and
\IEEEauthorblockN{Abhijith Ragav}
\IEEEauthorblockA{\textit{Georgia Institute of Technology}\\
Atlanta, USA\\
abhijithragav@gatech.edu}
}

\maketitle

\begin{abstract}
Ontologies and knowledge graphs require continuous evolution to remain accurate, but manual curation is labor intensive. Large Language Models (LLMs) possess vast unstructured knowledge yet struggle to maintain structured consistency. We propose Evo-DKD, a dual-decoder framework for autonomous ontology evolution that combines structured ontology traversal with unstructured text reasoning. Evo-DKD introduces two parallel decoding streams within an LLM: one generates candidate ontology edits (e.g., new concepts or relations) while the other produces natural-language justifications, with a dynamic attention-based gating mechanism deciding at each step how to blend structured and unstructured knowledge. Due to GPU constraints, we simulate dual-decoder behavior using prompt-based mode control in a single stream. The system operates in a closed reasoning loop: proposed edits are validated via consistency checks and cross-verification with the text explanations, then injected into the knowledge base, which informs subsequent reasoning. On healthcare ontology refinement, semantic search improvement, and cultural heritage timeline modeling, Evo-DKD outperforms structured-only and unstructured-only baselines in both precision of ontology updates and downstream task performance, with quantitative metrics and qualitative examples demonstrating the benefits of integrating structured and unstructured reasoning. Evo-DKD offers a new paradigm for LLM-driven knowledge base maintenance, combining symbolic and neural reasoning for sustainable ontology evolution.
\end{abstract}


\section{Introduction}
Modern AI systems increasingly rely on ontologies and knowledge graphs (KGs) to provide structured, verifiable knowledge for reasoning tasks. Ontologies formally define concepts and relationships within a domain, enabling consistent data integration and logical inference, but maintaining them is challenging: as domains evolve with new entities and concepts, ontologies require continuous updates, and manual curation or semi-automated pipelines struggle to keep pace with the rapid growth of unstructured information.

Large Language Models (LLMs) have demonstrated remarkable ability to absorb and generate knowledge from text, suggesting they could assist in extending ontologies, e.g., by populating ontology instances. However, naive application of LLMs often produces hallucinations that could corrupt a knowledge base, and LLM outputs lack the structured format required for direct ontology updates. The key challenge is aligning unstructured LLM reasoning with structured ontology editing in a reliable, automated loop.

In this work, we introduce Evo-DKD (Dual-Knowledge Decoding), a novel framework that tightly integrates an LLM's unstructured knowledge capabilities with structured ontology operations for autonomous ontology evolution. The core idea is a dual-decoder architecture: one stream produces structured outputs (ontology edits such as class insertions, relation assertions, or schema modifications), while the other produces unstructured outputs (natural language explanations and reasoning steps). These decoders operate in parallel, coordinated by a dynamic attention-based gating mechanism that determines each stream's contribution at every generation step, so that ontology edits are accompanied by grounded rationales, reducing unwarranted changes.

Our contributions are summarized as follows:
\begin{itemize}
    \item \textbf{Dual-Decoder Ontology Evolution Architecture:} A novel LLM architecture with two decoding streams (structured and unstructured) sharing a common encoder, with a gating module that dynamically balances their outputs—enabling simultaneous reasoning in natural language and formal ontology space.

    \item \textbf{Dynamic Gating and Validation Mechanism:} An attention-based gating router controls the interplay between decoders, and a validation module checks proposed edits using both the structured knowledge base (constraints, duplicates, inconsistencies) and the generated text (factual plausibility), ensuring only high-confidence edits enter the ontology.

    \item \textbf{Closed-Loop Autonomous Updating:} Each iteration's ontology updates are fed back into the LLM's context for the next iteration, so the knowledge base grows and improves over time without direct human input.

    \item \textbf{Experimental Evaluation:} Across three domains, a prompt-based simulation of Structured, Unstructured, and Full Dual-Decoder modes shows the Full Dual-Decoder consistently outperforming single-mode baselines on triple extraction accuracy (Exact and Relaxed Match), explanation quality (BLEU, BERTScore), and output reliability (LLM-Judge Score). We present cross-domain results, macro/micro score trends, and qualitative examples; while we do not implement a full dual-decoder architecture, the simulation demonstrates the value of integrating structured and unstructured reasoning, setting the stage for future work on gated decoding coordination.
\end{itemize}

The remainder of this paper is organized as follows. Section~\ref{Related Work} reviews related work on tool-augmented LLMs, ontology learning, and KG-to-text generation; Section~\ref{Architecture} details the Evo-DKD architecture, including the dual decoders, gating router, and validation/integration process; Section~\ref{Training} describes the training setup, metrics, and results; Section~\ref{Results} presents use cases with analysis; and Section~\ref{Discussion} discusses novelty, impact, limitations, and future directions.

\section{Related Work} \label{Related Work}

Recent advancements enhance LLMs through integration with external tools and knowledge bases to mitigate hallucinations and improve factual accuracy. Frameworks such as ReAct~\cite{yao2023react} and Toolformer~\cite{schick2023toolformer} enable LLMs to interleave reasoning steps with external API calls for retrieval and calculation. While effective, these approaches treat external knowledge sources as static reference points, with the LLM acting solely as a consumer rather than an active editor. In contrast, Evo-DKD positions the LLM as a contributor, proposing knowledge base edits through a dual-decoder architecture that ensures structured suggestions are consistently justified by unstructured textual reasoning.

Ontology evolution traditionally relies on manual or semi-automated pipelines that detect, propose, and validate ontology changes with significant human oversight~\cite{article}. Systems like NELL~\cite{10.5555/2898607.2898816} demonstrated continuous ontology updates through iterative web extraction pipelines, but rely on handcrafted modules rather than neural generation. LLMs4OL~\cite{10.1007/978-3-031-47240-4_22} evaluates zero-shot LLM prompts for ontology learning tasks, whereas OntoGenix~\cite{VALCALVO2025104042} integrates human review post-generation. Evo-DKD diverges by embedding validation within its decoding loop, autonomously updating ontologies in real time without external human intervention, while maintaining structured coherence through internal gating.

Knowledge graph-to-text and text-to-knowledge graph research offers bi-directional transformations between structured data and natural language. Graph-conditioned generation~\cite{gardent-etal-2017-creating}~\cite{su-etal-2021-plan-generate} verbalizes triples into coherent text, whereas neural relation extraction and prompt-based LLM methods~\cite{yao2025exploringlargelanguagemodels} convert text into structured triples. Evo-DKD uniquely combines both, producing structured ontology edits alongside textual explanations that mutually reinforce accuracy, enhancing reliability and facilitating continual autonomous ontology evolution.

\section{Architecture}
\label{Architecture}
\subsection{Dual-Knowledge Decoding Architecture}

Figure~\ref{fig:Architecture} provides an overview of the dual-decoder design and the closed-loop operation of the system. The model is built on a pre-trained Transformer-based LLM backbone, on top of which we instantiate two decoders specialized for different output modalities:

\begin{figure}[htbp]
    \centering
    \includegraphics[width=\columnwidth,height=0.65\columnwidth,keepaspectratio]{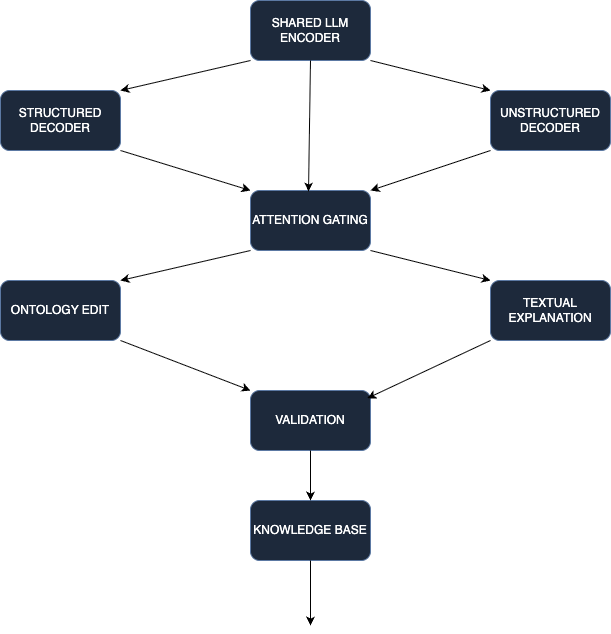}
    \caption{Evo-DKD dual-decoder architecture. The LLM backbone encodes the prompt with context from the current knowledge base. A Structured Decoder proposes an ontology edit while an Unstructured Decoder generates a textual explanation; an attention-based gating module dynamically balances the two streams at each generation step. Validated edits (checked for consistency and factual support) are applied to the Knowledge Base, updating context for the next iteration and closing the reasoning loop.}
    \label{fig:Architecture}
\end{figure}

\begin{itemize}
    \item \textbf{Structured Decoder (Ontology Stream):} Generates outputs in a structured format compatible with ontologies—triples like \texttt{(Diabetes, subclassOf, Disease)} or axioms in a formal language (OWL/RDF syntax)—with a vocabulary constrained to ontology elements. This decoder is responsible for proposing ontology edits.

    \item \textbf{Unstructured Decoder (Text Stream):} Generates free-form natural language to articulate reasoning, provide explanations or evidence, and narrate the proposed changes. This stream draws on the full vocabulary and knowledge encoded in the LLM to ensure that any structured proposal is contextually justified.
\end{itemize}

Both decoders receive input from the shared LLM encoder, which encodes the concatenation of (1) the user query or prompt, (2) the current state of the ontology, and (3) any retrieved or pre-loaded textual context (e.g., documents or facts relevant to the query or domain). In practice, the ontology state can be a set of embedding vectors or a textual serialization of key facts included in the prompt; the shared encoder gives both decoders a common understanding of input and context.

\subsection{Dual-Decoder Coordination via Dynamic Gating}
At a high level, the gating mechanism decides whether the model should next generate a structured element or elaborate in text, and whether the current structured proposal is consistent with the textual reasoning so far.

Concretely, at each decoding timestep \( t \), let \( h^{(s)}_t \) and \( h^{(u)}_t \) be the hidden states of the structured and unstructured decoders. The gating module takes these (and optionally the encoder context \( h^{(enc)} \)) to compute a weight \( \alpha_t \in [0,1] \) representing the emphasis on the structured decoder at that step (with \( 1-\alpha_t \) the emphasis on the text decoder):
\[
    \alpha_t = \sigma \bigl( W [h^{(s)}_t; h^{(u)}_t; h^{(enc)}] + b \bigr),
\]
where \( \sigma \) is a sigmoid; a softmax over multiple experts could generalize this, but here we have two. The decoders then receive a gating-adjusted context for the next token prediction. We recommend two strategies:

\paragraph{Switching Mode:} Treat the gating output in a hard binary way—if \( \alpha_t > 0.5 \), the model commits to a structured token at step \( t \); otherwise, it generates a word in the unstructured explanation, so the two decoders take turns producing tokens. Since uncontrolled switching could lead to jumbled output, switching is confined to boundaries of meaningful segments (e.g., complete triple vs. sentence).

\paragraph{Mixture Mode:} Use \( \alpha_t \) to form a convex combination of the two decoders' output distributions \( P^{(s)}_t \) and \( P^{(u)}_t \):
\[
    P^{(mix)}_t = \alpha_t \cdot P^{(s)}_t + (1-\alpha_t) \cdot P^{(u)}_t .
\]
Since the two vocabularies are disjoint, \( \alpha_t \) close to 1 yields a structured token and vice versa. This ``soft'' gating allows finer-grained control and smoother transitions.

The gating network's parameters are learned during fine-tuning on synthetic tasks where the desired balance is known (e.g., forcing alternating explanation text and ontology triples on a synthetic dataset of updates with rationales), or via reinforcement learning on end-task performance. Gating also enforces coherence between the streams: if the structured decoder proposes a relation that the text decoder has not justified—or contradicts—the gating mechanism can postpone finalizing that proposal, encouraging the text decoder to elaborate or revise. We note that the gating module is presented here as a design specification; in this work its behavior is approximated through prompt-based mode control (Section~\ref{Results}) rather than a trained gating network.

\subsection{Validation and Ontology Update Integration}
After decoding, Evo-DKD produces two outputs: (1) a candidate ontology edit from the structured decoder, and (2) a textual explanation from the unstructured decoder. Before applying the edit to the knowledge base, a validation step ensures the change is sound.

\subsubsection{Validation Module} This checks the structured proposal along two dimensions:

\textit{Ontology Consistency and Constraints:} The edit must not violate schema-level constraints (e.g., domain/range restrictions on a relation) and is checked for duplication or redundancy to avoid clutter. Edits failing these checks are rejected or flagged for manual review.

\textit{Justification Cross-Check:} We assess whether the unstructured explanation provides factual support for the edit, using the LLM itself (or a separate verification model) with a prompt such as: \textit{``Does the preceding explanation justify adding relationship R between A and B? Answer yes or no.''} If the justification is insufficient or contradictory, the edit is discarded or sent for further analysis.

\subsubsection{Ontology Update:} If validation passes, the new class, relation, or fact is inserted into the knowledge base, and the updated ontology is fed back into the next cycle: the encoder context then includes the newly added facts. Over many iterations, the ontology evolves.

Running continuously or on a schedule, Evo-DKD implements a closed-loop learning system: the LLM permanently records new inferred knowledge into the structured repository, improving future performance on queries requiring that knowledge (demonstrated in Section~\ref{RAG} with RAG), and each edit's stored explanation provides explainability that can be consulted later.

\section{Training} \label{Training}
\subsection{Modeling}
We employed TinyLlama-1.1B, chosen for feasibility under constrained GPU resources, fine-tuned for structured knowledge extraction across three domains: Healthcare (clinical statements such as drug-disease interactions), Semantic Search (user intent from queries), and Cultural Heritage (historical events and relationships between artifacts and figures).

The fine-tuning dataset comprised 600 synthetic examples containing diverse structured triples (subject, relation, object) with textual explanations justifying each assertion:
\begin{itemize}
\item{Healthcare: (Metformin, treats, Type 2 Diabetes) — "Metformin is a first-line therapy shown to manage blood glucose in patients with Type 2 Diabetes."}

\item{Semantic Search: (Users, searchFor, battery life) — "Search logs indicate that queries about battery life are common among smartphone users concerned with device longevity."}

\item{Cultural Heritage: (Rosetta Stone, enabledTranslationOf, ancient Egyptian scripts) — "The Rosetta Stone provided the key to deciphering ancient Egyptian scripts through its trilingual inscriptions."}
\end{itemize}

The model was fine-tuned using Hugging Face's Transformers library with the AdamW optimizer over three epochs (720 training steps), using gradient accumulation (4 steps), a learning rate of 2e-5, and linear warm-up for 50 steps. Our implementation uses a single-decoder approach, simulating dual-decoder functionality via chat-based prompting, and jointly predicts structured triples and explanations using a single language modeling loss over the combined sequence.

\subsection{Evaluation Metrics}
We evaluated the fine-tuned model using: \textbf{Precision} (fraction of proposed edits deemed correct); \textbf{Recall} (fraction of relevant changes successfully added); \textbf{Triple Extraction Accuracy} under both \textit{Exact Match} and \textit{Relaxed Match} criteria against ground-truth triples; \textbf{BLEU}, assessing quality and fluency of generated explanations against references; \textbf{BERTScore}, capturing semantic similarity and contextual accuracy of explanations; and an \textbf{LLM-Judge Score}, produced by a DistilBERT classifier (\texttt{\path{lvwerra/distilbert-imdb}}) as a lightweight automated proxy for explanation quality; stronger LLM-as-judge protocols or human evaluation are left to future work.

\subsection{Results Summary}
The fine-tuned \texttt{TinyLlama-1.1B} demonstrated good performance on test data across all evaluation metrics (Table~\ref{tab:metrics}), accurately predicting structured knowledge with coherent, human-like explanations. Such high precision and recall, combined with strong explanatory capability, suggests the suitability of our method for ontology refinement and semantic enrichment across diverse domains.

\begin{table}[h]
    \centering
    \begin{tabular}{lc}
        \hline
        \textbf{Metric} & \textbf{Score} \\
        \hline
        Relaxed Accuracy & 0.97 \\
        Exact Accuracy & 0.93 \\
        Precision & 0.98 \\
        Recall & 0.93 \\
        F1-Score & 0.95 \\
        BLEU Score for Explanations & 0.81 \\
        BERTScore & 0.88 \\
        LLM-Judge Score & 0.76 \\
        \hline
    \end{tabular}
    \caption{Performance Metrics}
    \label{tab:metrics}
\end{table}

\section{Results}\label{Results}
\subsection{Experimental Setup}
To simulate a dual-decoder architecture under constrained training resources, we designed three inference modes that isolate and combine the capabilities of structured and unstructured decoding:

\begin{itemize}
\item \textbf{Structured-only:} Extracts structured triples without natural language justifications, emulating traditional ontology population pipelines focused purely on formal knowledge representation.

\item \textbf{Unstructured-only:} Generates free-text explanations without structured outputs, mirroring text-only reasoning systems where facts remain implicit and require post-processing.

\item \textbf{Full Dual-Decoder:} Prompts for structured triples followed by explanatory text, reflecting the coordinated reasoning behavior of Evo-DKD and enabling assessment of how joint outputs support ontology evolution.
\end{itemize}

We used the fine-tuned TinyLlama-1.1B from Section~\ref{Training} and varied only the prompting strategy to emulate the three modes, without changing model weights or architecture. For each domain, we created evaluation sets of 40 carefully curated input-output pairs (120 total) and systematically computed the metrics above, allowing comparison of isolated versus integrated reasoning strategies across diverse ontology types.

\subsection{Evaluation Metrics and Results}
\subsubsection{Triple Relaxed Match Accuracy:}
\begin{figure}[t]
  \centering
  \begin{subfigure}{0.48\columnwidth}
    \centering
    \includegraphics[width=\columnwidth,height=0.42\columnwidth,keepaspectratio]{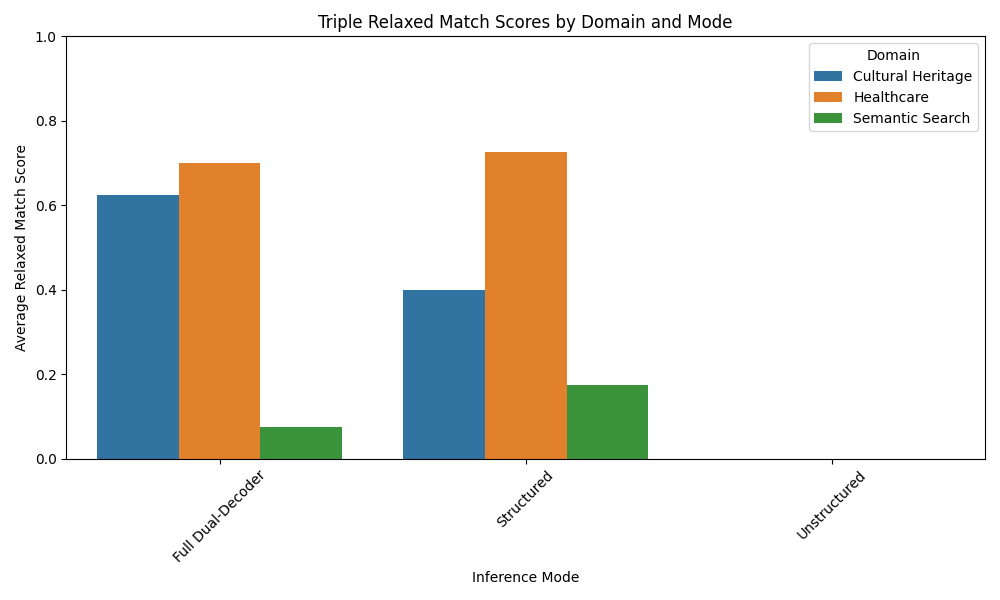}
  \end{subfigure}
  \hfill
  \begin{subfigure}{0.48\columnwidth}
    \centering
    \includegraphics[width=\columnwidth,height=0.42\columnwidth,keepaspectratio]{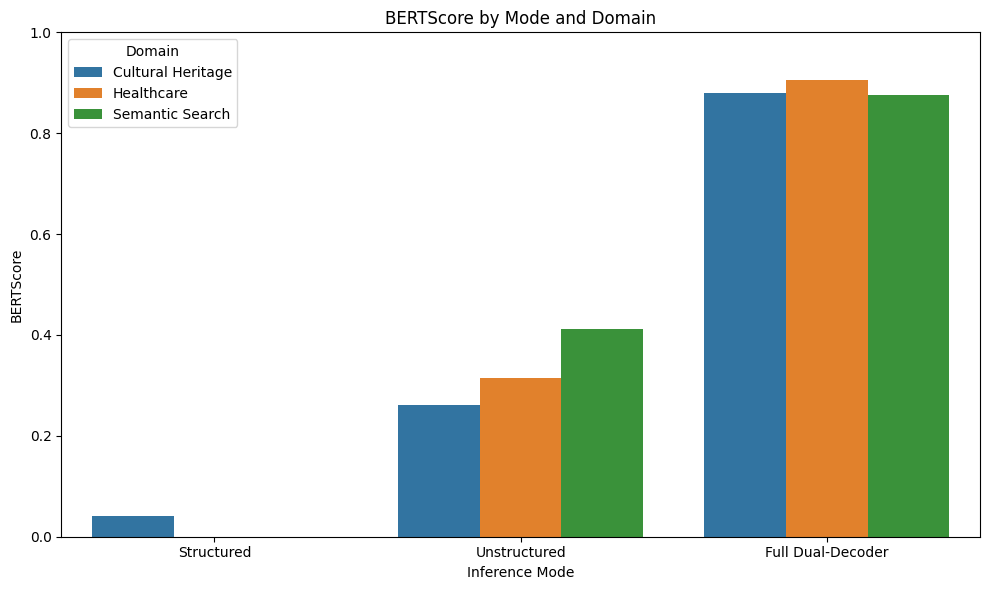}
  \end{subfigure}
  \caption{(a) Triple Extraction Accuracy (b) BERTScore}
  \label{fig:triple_and_bertscore}
  \label{fig:relaxed_match}
  \label{fig:bertscore}
\end{figure}

The Relaxed Match results (Figure~\ref{fig:relaxed_match}a) emphasize the benefits of the Full Dual-Decoder, particularly where surface-level variability in entity naming and relation phrasing is common. In Healthcare, both the Full Dual-Decoder and Structured-only modes achieved high relaxed match scores (0.7+), underscoring the model's ability to capture semantically equivalent relations phrased differently. In Cultural Heritage, the Full Dual-Decoder again clearly improved over Structured-only, suggesting that combining unstructured reasoning with structured outputs helps generalize across less standardized historical expressions and relation types. Semantic Search remained the most difficult domain across all modes, highlighting the difficulty of turning vague or abstract search-related phrases into clear, structured knowledge. Scores are 0 for Unstructured-only because it generates no triples to evaluate.

\subsubsection{Explanation Quality (BERTScore and BLEU:)}
\begin{figure}[t]
  \centering
  \begin{subfigure}{0.48\columnwidth}
    \centering
    \includegraphics[width=\columnwidth,height=0.42\columnwidth,keepaspectratio]{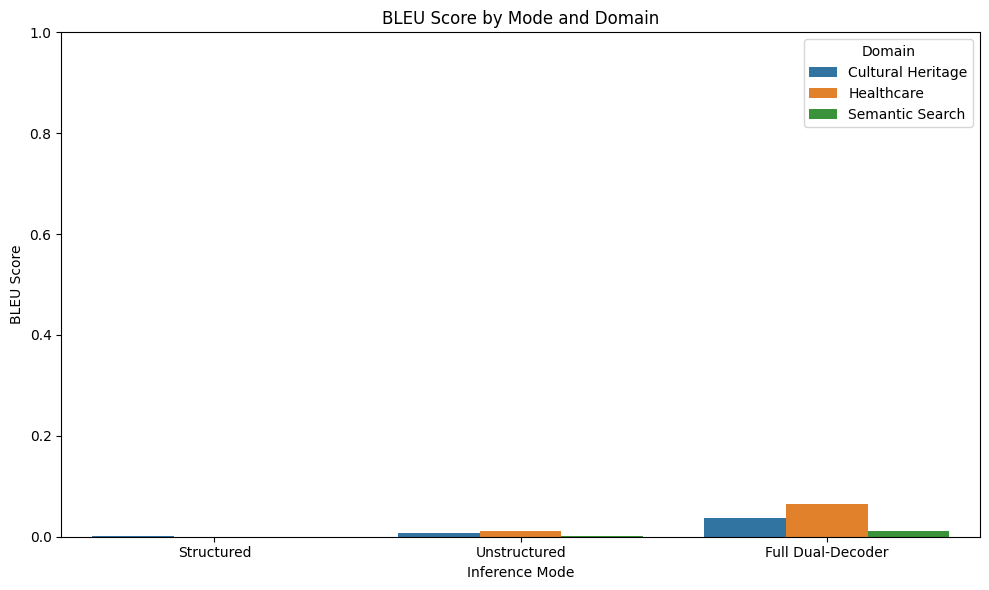}
  \end{subfigure}
  \hfill
  \begin{subfigure}{0.48\columnwidth}
    \centering
    \includegraphics[width=\columnwidth,height=0.42\columnwidth,keepaspectratio]{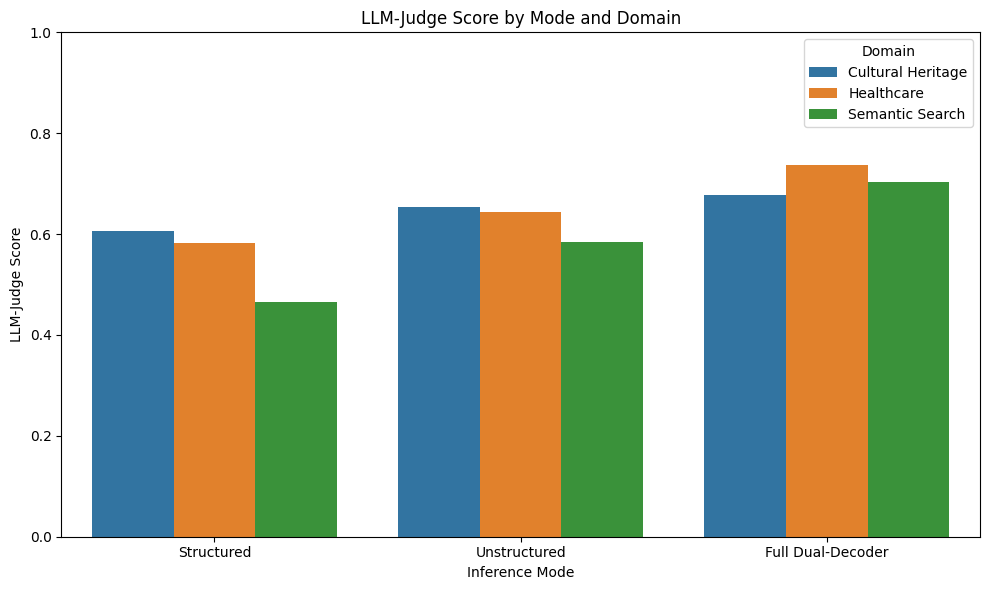}
  \end{subfigure}
  \caption{(a) BLEU Score (b) LLM Judge Score}
  \label{fig:bleu_and_judge}
  \label{fig:bleu}
  \label{fig:llm_judge}
\end{figure}

The BERTScore distribution (Figure~\ref{fig:bertscore}b) demonstrates superior semantic coherence in explanations from the \textbf{Full Dual-Decoder} mode, which substantially outperformed both single modes across all domains: its explanations align closely with references, underscoring the value of coordinated decoding. BLEU scores (Figure~\ref{fig:bleu}a), measuring textual similarity and human-likeness, were comparatively lower across modes, consistent with expected variability in human-generated explanations. Despite lower absolute values, this highlights the dual-decoder's flexibility in producing diverse, contextually rich textual outputs suited to real-world scenarios demanding nuanced justifications.

\subsubsection{Qualitative Evaluation (LLM-Judge Score:)}
LLM-Judge scores (Figure~\ref{fig:llm_judge}b) further validated the \textbf{Full Dual-Decoder} mode, which consistently scored higher than the other modes. This independent qualitative metric underscores its ability to generate accurate, compelling, and well-supported explanations, increasing practical trustworthiness.

\subsection{Cross-Domain Analysis}
\begin{figure}[t]
  \centering
  \begin{subfigure}{0.48\columnwidth}
    \centering
    \includegraphics[width=\columnwidth,height=0.42\columnwidth,keepaspectratio]{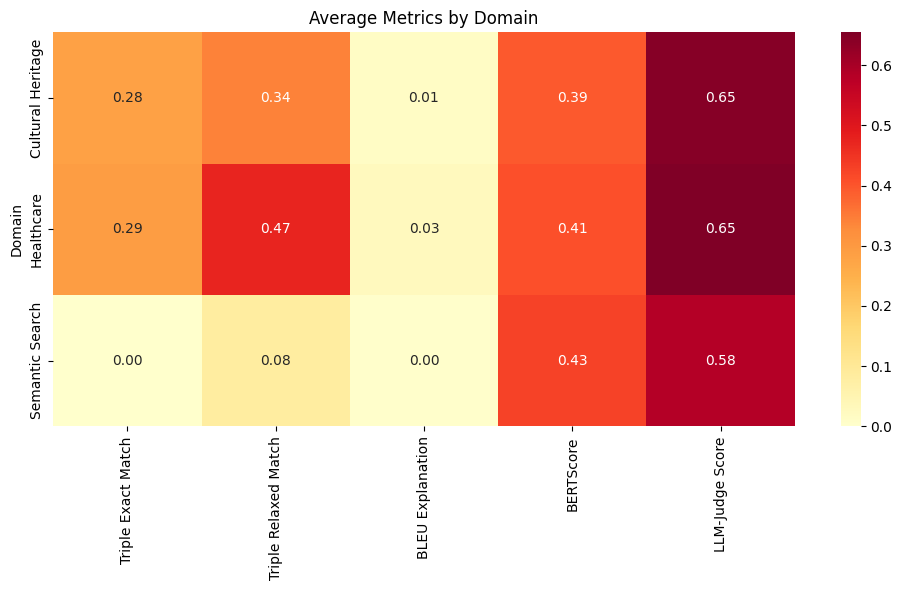}
  \end{subfigure}
  \hfill
  \begin{subfigure}{0.48\columnwidth}
    \centering
    \includegraphics[width=\columnwidth,height=0.42\columnwidth,keepaspectratio]{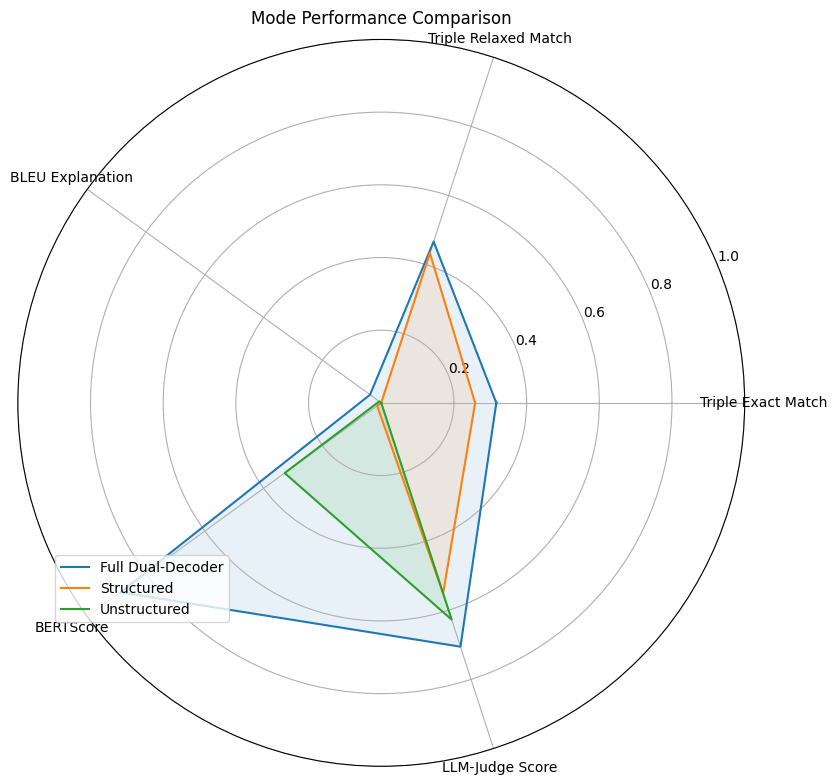}
  \end{subfigure}
  \caption{(a) Cross Domain Analysis (b) Overall Performance Analysis}
  \label{fig:crossdomain_and_overall}
  \label{fig:HeatMap}
  \label{fig:Figure7}
\end{figure}

The cross-domain heatmap (Figure~\ref{fig:HeatMap}a) revealed key differences across domains: \textbf{Healthcare} consistently exhibited high triple extraction performance, achieving the highest average Relaxed Match score, indicative of effective extraction of structured medical knowledge. \textbf{Cultural Heritage} showed intermediate performance, highlighting domain complexity but substantial gains under the Full Dual-Decoder mode. \textbf{Semantic Search} posed significant challenges, with lower average Exact and Relaxed Match metrics reflecting inherent domain ambiguities.

\subsection{Aggregate Trends Across Modes}
\begin{figure}[ht]
  \centering
\includegraphics[width=\columnwidth,height=0.45\columnwidth,keepaspectratio]{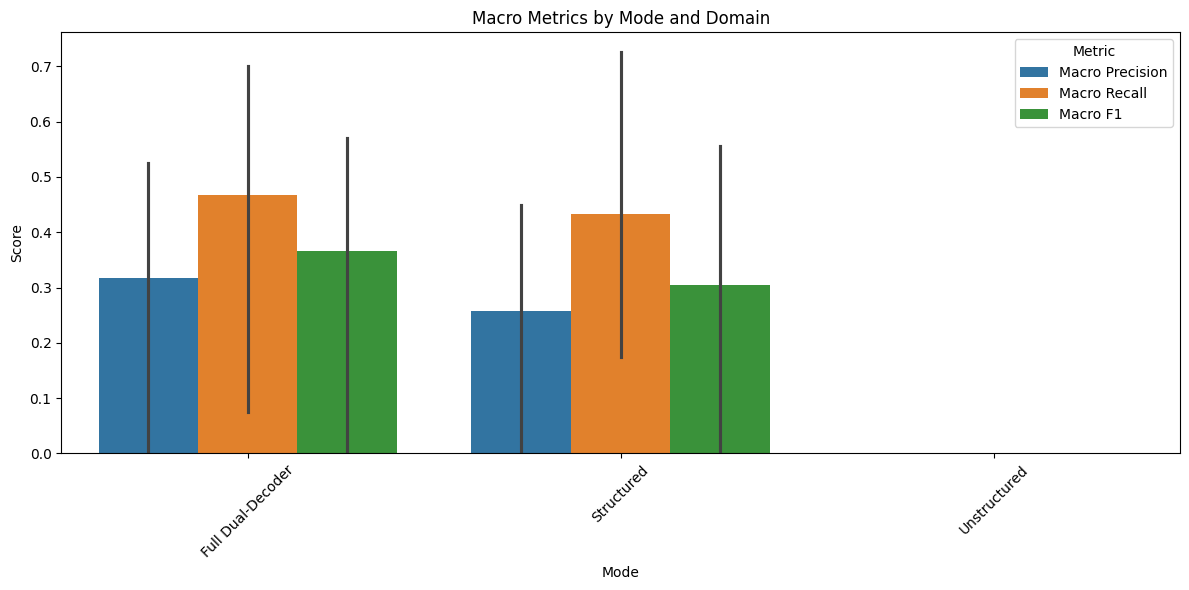}
  \caption{Macro metrics}
  \label{fig:Figure9}
\end{figure}

The radar chart (Figure~\ref{fig:Figure7}b) summarizes overall performance across inference modes. The \textbf{Full Dual-Decoder} distinctly outperformed single-decoder baselines by leveraging complementary strengths: Structured-only provided precise triple extraction but lacked textual reasoning, while Unstructured-only generated coherent explanations but struggled with structured representation; the dual-decoder bridged these gaps with balanced performance across all metrics. Macro metrics (Figure~\ref{fig:Figure9}), aggregating precision, recall, and F1 across domains, confirmed balanced precision-recall trade-offs—crucial for accurate knowledge retrieval.

\subsection{Case Study: Qualitative Evaluation - Ontology Evolution in Healthcare} \label{RAG}

We qualitatively demonstrate Evo-DKD's capability for dynamic ontology evolution using \textbf{healthcare} as a representative domain. Initially, the knowledge graph contained foundational triples:

\begin{table}[h]
    \centering
    \begin{tabular}{ccc}
        \hline
        (Aspirin, & reducesRiskOf, & heart attacks) \\
        (Metformin, & treats, & Type 2 Diabetes) \\
        (Ibuprofen, & alleviates, & pain) \\
        \hline
    \end{tabular}
    \caption{Triplets}
\end{table}

Given the new input \textit{``Ozempic helps manage weight loss in diabetic patients''}, Evo-DKD generated the structured triple (\textit{Ozempic}, \textit{manages}, \textit{weight}) with the explanatory text:

\begin{quote}
``Doctors recommend Ozempic to help manage weight loss in diabetes.''
\end{quote}

Immediately integrating this triple into the knowledge graph significantly improved downstream retrieval-augmented generation (RAG), shown below for the user query \textit{``What drugs are used for weight loss in diabetes?''} Our lightweight RAG pipeline retrieves the top-k (k=2) most relevant KG facts and explanations using semantic similarity, compiles them into a structured context block, and passes them to a Gemini-powered LLM whose response is constrained to KG content—grounding answers in the currently stored triples and explanations for transparent, up-to-date reasoning.

\begin{figure}[t]
  \centering
  \begin{subfigure}{0.48\columnwidth}
    \centering
    \includegraphics[width=\columnwidth,height=0.52\columnwidth,keepaspectratio]{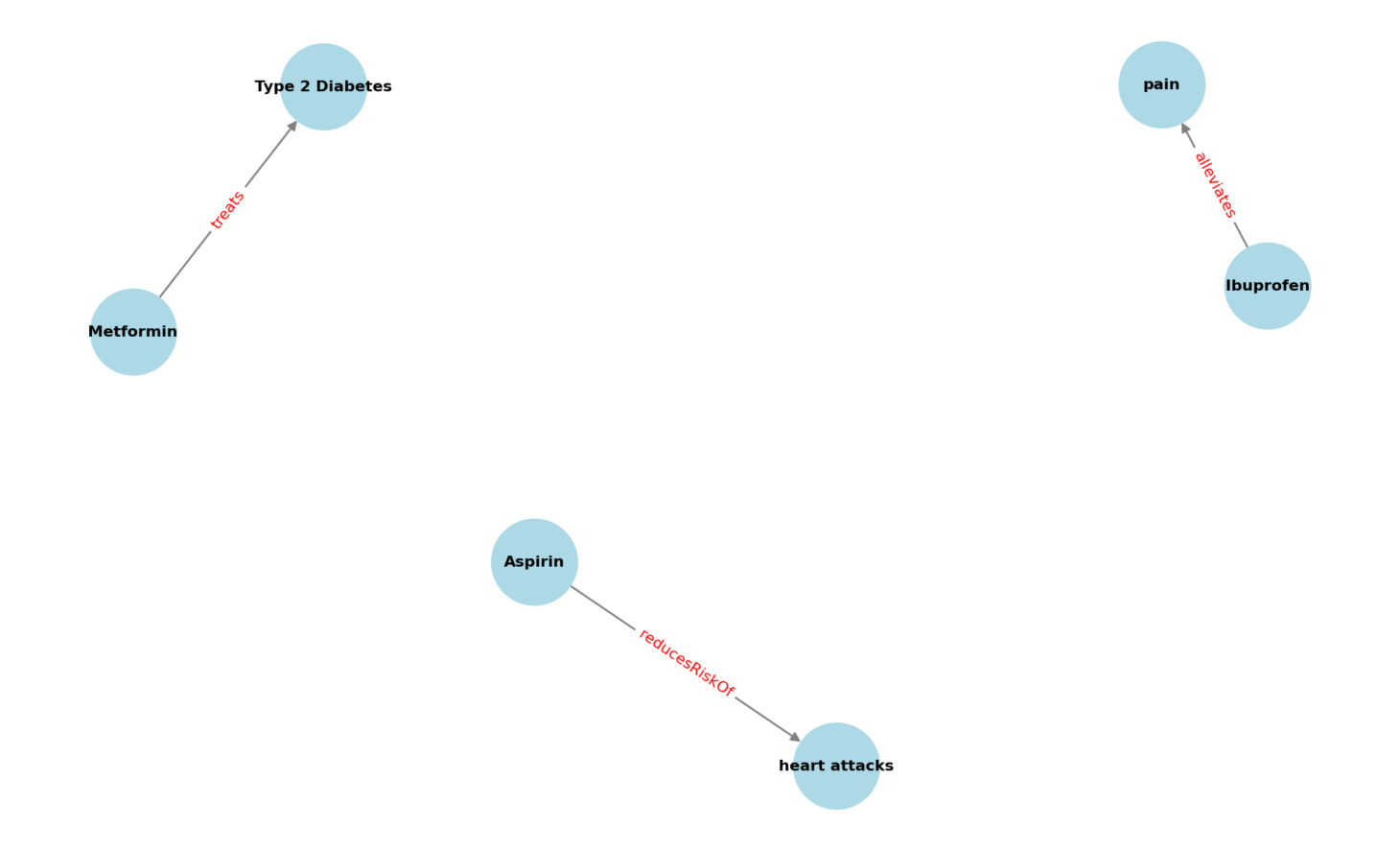}
  \end{subfigure}
  \hfill
  \begin{subfigure}{0.48\columnwidth}
    \centering
    \includegraphics[width=\columnwidth,height=0.52\columnwidth,keepaspectratio]{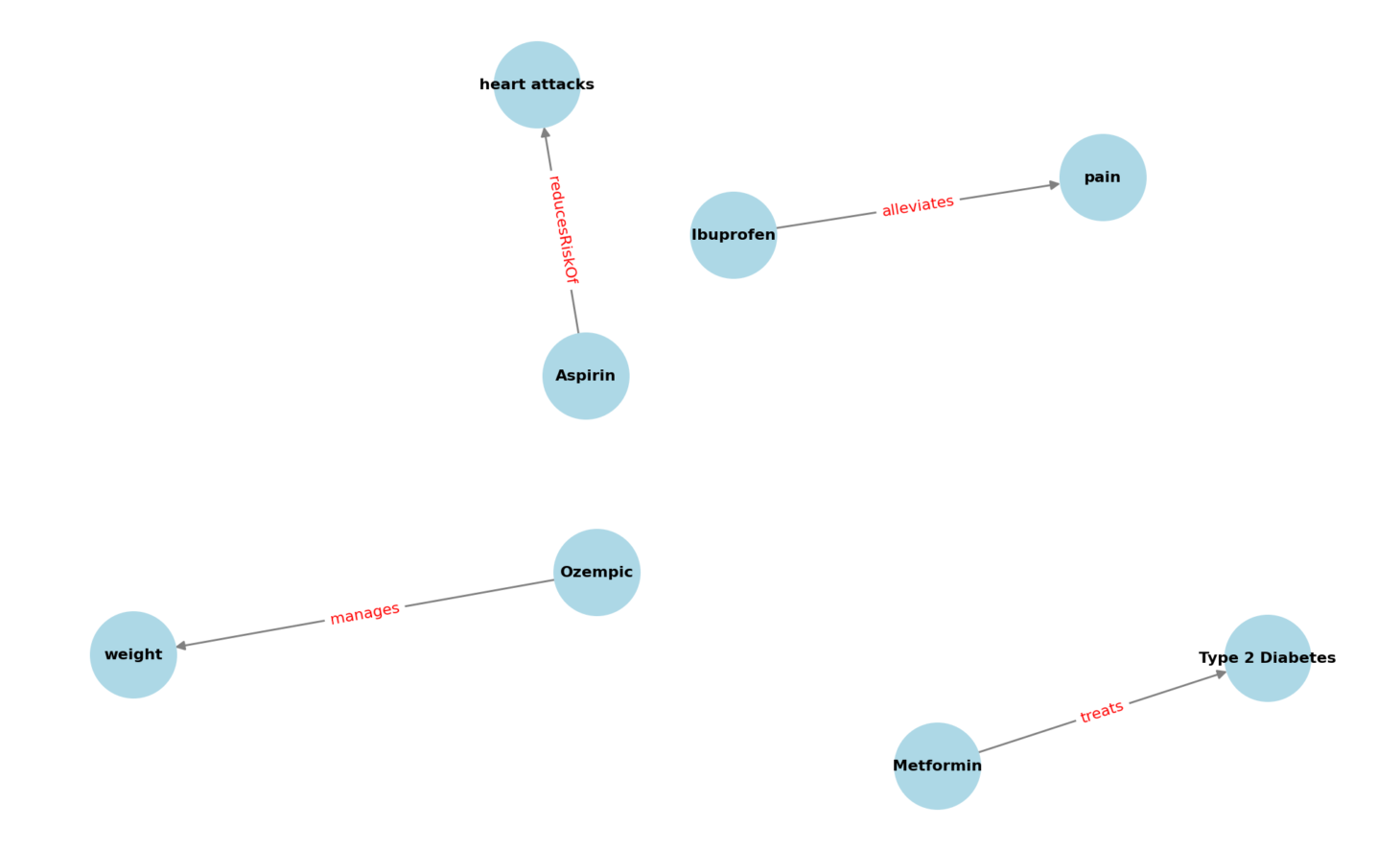}
  \end{subfigure}
  \caption{(a) Knowledge Graph with  foundational triples (b)  Knowledge Graph after user input}
  \label{fig:kg_before_after}
  \label{fig:KG_BEFORE}
  \label{fig:KG_AFTER}
\end{figure}

\subsubsection{RAG answer with Base Knowledge Graph (Figure~\ref{fig:KG_BEFORE}a):}

The user query yielded no relevant information: \textit{``The provided facts do not contain information about drugs used for weight loss in diabetes.''}

\subsubsection{RAG answer with Updated Knowledge Graph (Figure~\ref{fig:KG_AFTER}b) with user input:}
The same query correctly identified \textit{Ozempic} as a recommended treatment—new fact added: (\textit{Ozempic}, \textit{manages}, \textit{weight}), with explanation \textit{``Doctors recommend Ozempic to help manage weight loss in diabetes.''}

Evo-DKD thus enables real-time ontology growth: triples extracted from user input are immediately injected into the KG, improving downstream QA via RAG—a dynamic loop with practical, deployable benefits.

\section{Discussion}\label{Discussion}
\subsection{Novelty and Contributions}

Evo-DKD represents a significant step toward neuro-symbolic learning in LLMs: combining symbolic output (structured ontology edits) with neural output (language) in one model goes beyond prior retrieval-augmented or tool-using approaches. While dual-decoder setups have been explored for guided text generation or multi-task learning, using them for simultaneous knowledge graph updates and explanatory text is new, enabling LLM self-reflection—the model must ``convince itself'' of a fact before adding it.

The dynamic gating mechanism acts as a learnable mixture-of-experts choosing between a ``text expert'' and ``symbol expert.'' Unlike static mixtures, it is context-sensitive at the token level and could extend to additional experts (e.g., other modalities or knowledge formats).

Another novelty is \textbf{closed-loop training}: Evo-DKD updates its own knowledge while ingesting new data, blurring training and inference—a step toward continual/lifelong learning. Traditional LLMs are fixed post-training and rely on external retrieval; Evo-DKD suggests a path to internalize updates structurally.

\subsection{Impact and Applications}

Autonomous knowledge base updating has broad implications: in enterprises, dramatically reducing the cost of maintaining up-to-date knowledge graphs, ontologies, or databases; in science, keeping domain ontologies current with the latest published findings; in personal assistants, accumulating user preferences in structured memory for personalization.

The dual outputs also make the system more interpretable: each addition carries an auditable explanation, improving trust in AI-curated knowledge bases.

\subsection{Limitations}

Despite its promise, Evo-DKD has limitations:

\begin{itemize}
\item \textbf{Reliance on LLM's Internal Knowledge:} The unstructured decoder's output is only as good as the LLM's knowledge; if the model lacks domain knowledge or the context misses key evidence, Evo-DKD may fail to add important facts (false negatives). \textbf{Mitigation:} integrate external retrieval, e.g., querying a search engine when uncertain.

\item \textbf{Complex Edits and Ontology Restructuring:} Our work focuses on fact additions or simple changes; complex ontology refactoring (e.g., removing sub-hierarchies or merging duplicate concepts) requires global planning and multi-step reasoning beyond our one-pass approach, left for future work via multi-turn dialogues or planning decoders.

\item \textbf{Validation Independence and Scale:} Self-verification by the same model is not fully independent validation; integrating formal ontology reasoners and human oversight is future work. Moreover, our evaluation uses small synthetic datasets (600 training / 120 test examples), so results should be read as a controlled simulation study rather than evidence at scale.
\end{itemize}

\subsection{Future Work}

One avenue for future research is multi-modality, e.g., images or tables as context: an art ontology could benefit from an image decoder recognizing paintings and suggesting metadata alongside a text decoder reading descriptions. Another is user-in-the-loop feedback: asking clarifying questions when uncertain, as interactive agents do.

Finally, future work will train dedicated structured and unstructured decoders with dynamic gating—exploring why decoder-level coordination improves factual consistency—enabling finer-grained control across knowledge-rich domains and potentially improving BLEU through more fluent explanations rather than hand-crafted prompts.

\section{Conclusion}
We presented Evo-DKD, a dual-knowledge decoding framework that empowers LLMs to autonomously evolve ontologies. Structured and unstructured decoding, coordinated through a dynamic attention-based gating mechanism, lets the system propose and validate new knowledge in context, maintaining alignment between LLM outputs and an evolving knowledge base. Our implementation simulates dual-decoder dynamics through controlled prompting that guides generation along distinct structured and textual reasoning paths, capturing complementary behaviors within a single decoder. Evaluations across healthcare, semantic search, and cultural heritage show consistent improvements over single-modality baselines, with high-quality ontology updates and measurable downstream gains. These results position Evo-DKD as a prototype of self-improving knowledge systems, laying groundwork for language models that not only consume knowledge but also curate, update, and justify it.

\balance
\bibliographystyle{IEEEtran}
\bibliography{references}

@inproceedings{
yao2023react,
title={ReAct: Synergizing Reasoning and Acting in Language Models},
author={Shunyu Yao and Jeffrey Zhao and Dian Yu and Nan Du and Izhak Shafran and Karthik R Narasimhan and Yuan Cao},
booktitle={The Eleventh International Conference on Learning Representations },
year={2023},
url={https://openreview.net/forum?id=WE_vluYUL-X}
}

@inproceedings{
schick2023toolformer,
title={Toolformer: Language Models Can Teach Themselves to Use Tools},
author={Timo Schick and Jane Dwivedi-Yu and Roberto Dessi and Roberta Raileanu and Maria Lomeli and Eric Hambro and Luke Zettlemoyer and Nicola Cancedda and Thomas Scialom},
booktitle={Thirty-seventh Conference on Neural Information Processing Systems},
year={2023},
url={https://openreview.net/forum?id=Yacmpz84TH}
}

@inproceedings{10.5555/2898607.2898816,
author = {Carlson, Andrew and Betteridge, Justin and Kisiel, Bryan and Settles, Burr and Hruschka, Estevam R. and Mitchell, Tom M.},
title = {Toward an architecture for never-ending language learning},
year = {2010},
publisher = {AAAI Press},
booktitle = {Proceedings of the Twenty-Fourth AAAI Conference on Artificial Intelligence},
pages = {1306–1313},
numpages = {8},
location = {Atlanta, Georgia},
series = {AAAI'10}
}

@article{article,
author = {Zablith, Fouad and Antoniou, Grigoris and d'Aquin, Mathieu and Flouris, Giorgos and Kondylakis, Haridimos and Motta, Enrico and Plexousakis, Dimitris and Sabou, Marta},
year = {2015},
month = {01},
pages = {45-75},
title = {Ontology evolution: A process-centric survey},
volume = {30},
journal = {The Knowledge Engineering Review},
doi = {10.1017/S0269888913000349}
}

@inproceedings{gardent-etal-2017-creating,
    title = "Creating Training Corpora for {NLG} Micro-Planners",
    author = "Gardent, Claire  and
      Shimorina, Anastasia  and
      Narayan, Shashi  and
      Perez-Beltrachini, Laura",
    editor = "Barzilay, Regina  and
      Kan, Min-Yen",
    booktitle = "Proceedings of the 55th Annual Meeting of the Association for Computational Linguistics (Volume 1: Long Papers)",
    month = jul,
    year = "2017",
    address = "Vancouver, Canada",
    publisher = "Association for Computational Linguistics",
    url = "https://aclanthology.org/P17-1017/",
    doi = "10.18653/v1/P17-1017",
    pages = "179--188"
}

@inproceedings{su-etal-2021-plan-generate,
    title = "Plan-then-Generate: Controlled Data-to-Text Generation via Planning",
    author = "Su, Yixuan  and
      Vandyke, David  and
      Wang, Sihui  and
      Fang, Yimai  and
      Collier, Nigel",
    editor = "Moens, Marie-Francine  and
      Huang, Xuanjing  and
      Specia, Lucia  and
      Yih, Scott Wen-tau",
    booktitle = "Findings of the Association for Computational Linguistics: EMNLP 2021",
    month = nov,
    year = "2021",
    address = "Punta Cana, Dominican Republic",
    publisher = "Association for Computational Linguistics",
    url = "https://aclanthology.org/2021.findings-emnlp.76/",
    doi = "10.18653/v1/2021.findings-emnlp.76",
    pages = "895--909"
}

@inproceedings{10.1007/978-3-031-47240-4_22,
author = {Babaei Giglou, Hamed and D’Souza, Jennifer and Auer, S\"{o}ren},
title = {LLMs4OL: Large Language Models for~Ontology Learning},
year = {2023},
isbn = {978-3-031-47239-8},
publisher = {Springer-Verlag},
address = {Berlin, Heidelberg},
url = {https://doi.org/10.1007/978-3-031-47240-4_22},
doi = {10.1007/978-3-031-47240-4_22},
booktitle = {The Semantic Web – ISWC 2023: 22nd International Semantic Web Conference, Athens, Greece, November 6–10, 2023, Proceedings, Part I},
pages = {408–427},
numpages = {20},
location = {Athens, Greece}
}

@article{VALCALVO2025104042,
title = {OntoGenix: Leveraging Large Language Models for enhanced ontology engineering from datasets},
journal = {Information Processing \& Management},
volume = {62},
number = {3},
pages = {104042},
year = {2025},
issn = {0306-4573},
doi = {https://doi.org/10.1016/j.ipm.2024.104042},
url = {https://www.sciencedirect.com/science/article/pii/S0306457324004011},
author = {Mikel Val-Calvo and Mikel {Egaña Aranguren} and Juan Mulero-Hernández and Ginés Almagro-Hernández and Prashant Deshmukh and José Antonio Bernabé-Díaz and Paola Espinoza-Arias and José Luis Sánchez-Fernández and Juergen Mueller and Jesualdo Tomás Fernández-Breis}
}

@misc{yao2025exploringlargelanguagemodels,
      title={Exploring Large Language Models for Knowledge Graph Completion}, 
      author={Liang Yao and Jiazhen Peng and Chengsheng Mao and Yuan Luo},
      year={2025},
      eprint={2308.13916},
      archivePrefix={arXiv},
      primaryClass={cs.CL},
      url={https://arxiv.org/abs/2308.13916}, 
}

\end{document}